%% file: paper.tex
\documentclass[runningheads]{llncs}

\usepackage[T1]{fontenc}
\usepackage{graphicx}

\usepackage{fancyhdr}
\fancypagestyle{plain}{%
  \fancyhf{}%
  \fancyfoot[L]{\small International Semantic Intelligence Conference (ISIC), Germany, 2025}%
  \fancyfoot[R]{\thepage}%
}
\usepackage{cite}
\usepackage{amsmath,amssymb,amsfonts}
\usepackage{algorithmic}
\usepackage{graphicx}
\usepackage{textcomp}
\usepackage{xcolor}
\usepackage{soul}
\usepackage{xcolor}
\usepackage{booktabs}

\usepackage{listings}
\usepackage[dvipsnames,svgnames,x11names,table]{xcolor}

\usepackage{comment}
\usepackage[pdfstartview={XYZ null null 1.2}]{hyperref}
\usepackage[numbers]{natbib}
\def\BibTeX{{\rm B\kern-.05em{\sc i\kern-.025em b}\kern-.08em
    T\kern-.1667em\lower.7ex\hbox{E}\kern-.125emX}}

\usepackage{xurl}
\usepackage{hyperref}
\begin{document}

\title{Interpretable Question Answering with Knowledge Graphs}

\author{
Kartikeya Aneja\inst{1,3}
\and
Manasvi Srivastava\inst{2,3}
\and 
Subhayan Das\inst{4}
\and
Nagender Aneja\inst{5}
}
\authorrunning{K. Aneja et al.}

\institute{
Department of Electrical and Computer Engineering, University of Wisconsin-Madison, Madison, USA \email{kaneja@wisc.edu}
\and
Department of Computer Science, Banasthali Vidyapeeth, Rajasthan, India \email{manasvisrivastava26@gmail.com}
\and
Ansyst Consulting, Gurgaon, India
\and
Huawei Ireland Research Center, Dublin, Ireland \email{subhayan.das@h-partners.com}
\and
Bradley Department of Electrical and Computer Engineering, Virginia Tech, Blacksburg, VA, USA \email{naneja@vt.edu}
}
\maketitle              

\thispagestyle{plain}

\begin{abstract}
This paper presents a question answering system that operates exclusively on a knowledge graph retrieval without relying on retrieval augmented generation (RAG) with large language models (LLMs). Instead, a small paraphraser model is used to paraphrase the entity relationship edges retrieved from querying the knowledge graph. The proposed pipeline is divided into two main stages. The first stage involves pre-processing a document to generate sets of question-answer (QA) pairs. The second stage converts these QAs into a knowledge graph from which graph-based retrieval is performed using embeddings and fuzzy techniques. The graph is queried, re-ranked, and paraphrased to generate a final answer. This work includes an evaluation using LLM-as-a-judge on the CRAG benchmark, which resulted in accuracies of 71.9\% and 54.4\% using LLAMA-3.2 and GPT-3.5-Turbo, respectively.

\keywords{Knowledge Graph  \and Question Answering \and RAG}
\end{abstract}

\input{text}

\bibliographystyle{IEEEtranN}
\bibliography{refs}

\end{document}

%% file: text.tex
\section{INTRODUCTION}
Question Answering (QA) systems aim to provide precise and contextually relevant answers to user queries, often using structured or unstructured knowledge sources. Recently, there has been interest in Retrieval-Augmented Generation (RAG)\cite{lewis2020retrieval}, where a user's query retrieves relevant text chunks, which are then passed to an LLM for answer generation. However, these systems are heavily reliant on unstructured documents and prone to hallucination \cite{xu2024hallucination, ji2023survey} and have limited transparency.

In contrast, Knowledge Graphs (KGs) provide a structured representation of information in the form of entities (nodes) and their relationships (edges). The KGs can be created from both structured and unstructured data. This formalism supports interpretable reasoning and contextual association, making them particularly suitable for knowledge-intensive QA tasks. Traditional KG-based QA systems often require complex semantic parsing and rule-based components, limiting their scalability and adaptability.

This research presents a Knowledge Graph-based QA system that does not rely on traditional chunk retrieval using RAG but instead constructs and queries a structured knowledge graph generated from QA pairs. Unlike RAG frameworks that use text chunking and dense retrieval, this approach builds a graph-based abstraction of knowledge, enabling semantic and entity-based search across the document's contents \cite{edge2024local, larson2024graphragblog, peng2024graph}.

Our methodology consists of two primary phases. 

First, QA pairs are generated from documents using a prompt-driven language model. These pairs are then converted into a knowledge graph via entity-relation extraction using \textit{LLMGraphTransformer} by \textit{Langchain} with \textit{GPT-3.5-Turbo}. In the second stage, user queries are answered by retrieving relevant subgraphs based on embedding similarity and fuzzy matching. Retrieved information is passed through a lightweight paraphrasing model \textit{tuner007/pegasus\_paraphrase} to produce coherent, human-readable answers. We evaluated the performance using LLM-as-a-judge.

By combining structured KG retrieval with paraphrased natural language responses, the proposed research offers an interpretable alternative to traditional RAG pipelines. It demonstrates practical benefits in legal and technical domains where traceability and factual consistency are critical.

\begin{table*}[htbp!]
\centering
\caption{QA Graph-Based Answering Flow}
\label{table:QAGraph-BasedAnsweringFlow}
\renewcommand{\arraystretch}{1.5}
\begin{tabular}{|p{5cm} | p{7cm}|}
\hline
\textbf{Component} & \textbf{Description} \\
\hline
Document Input & PDF containing contractual clauses \\
QA Pairs & Extracted using prompt-based QA generation \\
Graph Creation & GPT-3.5 identifies entities, relations \\
Embeddings & text-embedding-3-large on nodes and types \\
Retrieval & Cosine and fuzzy search on entities \\
Reranker & BAAI/bge-reranker-large on triples \\
Answer & tuner007/pegasus\_paraphrase paraphrasing \\
\hline
\end{tabular}
\end{table*}

\section{Related Work}
This section explains related work on knowledge graph-based question answering while avoiding LLM-based RAG. Related work can be categorized into two broad approaches: (i) semantic parsing that converts a question into a logical query, e.g., Cypher in the case of Neo4j, and executes it on the KG, and (ii) retrieval/embedding methods that can find relevant subgraphs or entities and paths without building a full logical query.

\citet{gu2022knowledge} reviewed research on question answering over knowledge bases for semantic-parsing and proposed joint entity linking and parsing. Most semantic parsing systems target either SPARQL, a semantic query language for data stored in Resource Description Framework (RDF) format, or use Cypher, a standard query language for property graph databases. \citet{du2023relation} proposed constructing a dual relation graph where relations are nodes, and edges connect relations sharing head or tail entities. Their approach alternates between (1) reasoning on the primal entity graph, (2) propagating information on the dual relation graph, and (3) enabling interaction between the two graphs. 

\citet{zhao2022implicit} proposed linking relation phrases to relation paths instead of single relations. The authors introduce a path ranking model that aligns both textual (word embeddings) and structural (KG embeddings) information, utilizing a gated attention mechanism with external paraphrase dictionaries to handle vague phrases. \citet{yang2024crag} highlighted the limitations of existing RAG approaches, showing that even advanced LLMs with RAG achieve only 44–63\% accuracy on dynamic, diverse questions. Although CRAG was built to benchmark RAG, our KG-only pipeline attains competitive accuracy on CRAG.

The proposed system leverages structured KG construction from QA pairs, followed by embedding-based retrieval, fuzzy entity matching, reranking, and paraphrasing. This design emphasizes interpretability, traceability, and reduced hallucination, providing clearer reasoning paths compared to black-box RAG methods.

\section{RESEARCH METHODOLOGY}

\subsection{QA Generation from Document}
We used two datasets: one is a PDF document, and the other is the CRAG dataset \cite{yang2024crag}. For our PDF document, we generated question-answer (QA) pairs. The text from the PDF is first divided into semantically meaningful units using a rule-based segmentation strategy. This strategy leverages structural cues, such as paragraph breaks, bullet points, clause identifiers, and length thresholds, to ensure that each chunk represents a logically coherent segment. These segments serve as atomic knowledge blocks for downstream indexing and retrieval. For each extracted chunk, the Hugging Face model \textit{iarfmoose/t5-base-question-generator} is used to generate a list of relevant question-answer pairs automatically. These pairs are designed to extract atomic facts and knowledge from the text and are stored in JSON format for further processing. In the CRAG, there are two tasks with 2706 question-answer pairs.

\subsection{Knowledge Graph Creation}
This section describes the methodology for constructing a Knowledge Graph from question–answer (QA) pairs and the subsequent process of querying it for relevant information. Given an input question, the system identifies and retrieves the associated entities and relationships from the graph. These retrieved entities and relationships are then fed into a lightweight paraphrasing model, which synthesizes the final answer in natural language. Table~\ref{table:QAGraph-BasedAnsweringFlow} presents a high-level overview of the Knowledge Graph-based answer generation workflow.

\subsubsection{Knowledge Graph Construction} 

Sets of question–answer (QA) pairs are processed through the LLMGraphTransformer of LangChain, utilizing the GPT-3.5 Turbo language model to extract entities and their semantic relationships. The additional information to consider Clauses and Numerical Values as nodes was passed to the LLMGraphTransformer as a prompt shown in Fig. \ref{lst:llm_transformer}. For this research, sets of twenty QA pairs are passed to LLMGraphTransformer at a time.

\begin{figure}
\centering
\begin{lstlisting}[language=Python]
prompt = "Clauses and Numerical Values, etc count as nodes"
llm_transformer = LLMGraphTransformer(llm=llm, additional_instructions=prompt)
\end{lstlisting}
\caption{LLM Graph Transformer setup}
\label{lst:llm_transformer}
\end{figure}

These nodes and relationships are used to construct a Knowledge Graph, which is stored in a Neo4j graph database. After processing all QA pairs, the resulting Knowledge Graph encapsulates the document's knowledge in a structured form. In this graph, nodes represent entities, while edges denote the relationships between them. We created two knowledge graphs, one for each dataset.

We then generated embeddings for each node and node type using transformer-based sentence embeddings via the \textit{text-embedding-3-large} model. These embeddings are computed internally by Neo4j using the specified model and subsequently stored within the graph database. In addition to individual node embeddings, we also computed and stored embeddings for node types. Embeddings for node types are also important, as nodes represent specific instances (e.g., individual people), which may be less informative for answering general or class-level questions—such as those referring to a group or category rather than a single entity.

\subsection{Retrieval Phase}
The retrieval phase consists of three key steps:

\textit{Node-Level Semantic Matching}: We first compute the cosine similarity between the embedding of the input question and the embeddings of candidate nodes in the Knowledge Graph. Nodes with the highest similarity scores are selected along with their respective immediate relationships.

\textit{Type-Level Generalization Retrieval}: In the second step, we identify the node type that is most semantically similar to the input question based on type embeddings. Once the top-matching node type is found, all nodes and relationships associated with that type are selected. This enables the retrieval of more generalized or abstract knowledge, especially useful when the question refers to a category (e.g., "scientists") rather than a specific instance.

\textit{Fuzzy Entity Matching}: Finally, we identify entities explicitly mentioned in the input question using a lightweight Hugging Face model \textit{dslim/bert-base-NER}. Then fuzzy matching is performed against the graph nodes. We allow an edit distance of up to three, enabling the system to account for minor variations or misspellings in entity names during retrieval.

\subsection{Paraphrase Phase}
Following the retrieval, the selected nodes and their associated relationships are passed to a lightweight paraphrasing model. This model generates a fluent natural language response by synthesizing the retrieved information into a coherent and contextually appropriate answer. The \textit{tuner007/pegasus\_paraphrase} is used for paraphrasing, and experiments suggest that performance can be improved with better-trained paraphraser models.

\subsection{Reranker Phase}
The paraphrased answers are subsequently passed to a HuggingFace reranking model(\textit{BAAI/bge-reranker -large}), which ranks them based on their semantic relevance to the input question. We also experimented with the reranker model before the Paraphrase phase and found no significant difference in the accuracy. The top five highest-ranked responses are selected as the final answers for evaluation.

\section{EVALUATION METHODOLOGY AND RESULTS}
This section presents the evaluation results obtained using automated assessment via an LLM-as-a-judge \cite{zheng2023judging}.

\subsection{LLM-as-a-Judge Evaluation}
We employed two large language models, Llama-3.2 and GPT-3.5 Turbo, as automatic evaluators to assess the correctness of the generated answers, as shown in Fig. \ref{lst:llm_evaluator}. Specifically, for each original question used in constructing the Knowledge Graph, we prompted the LLM to determine whether the ground-truth answer was present among the top five candidate answers produced by our system. We used a lenient prompt because a lightweight paraphraser was used in the answer generation, which can sometimes result in some broken English. Using this approach, the system achieved an accuracy of $89.6\%$ by Llama-3.2 and 78.3\% by gpt-3.5-turbo for our PDF dataset. The high accuracy shows that the retrieval from the knowledge graph performed well. 

For the CRAG dataset, we had 2706 QA pairs; however, 308 pairs had answer fields marked as "invalid question." Thus, valid questions were 2398. The accuracy of the CRAG dataset is 71.9\% by Llama-3.2 and 54.4\% by GPT-3.5-turbo. This is in comparison to the reported accuracy of 63\% \cite{yang2024crag}. We also analyzed a sample of incorrect questions and found that entity mismatch was the primary reason. The experiments demonstrate that enhancing entity-relationship extraction can further enhance performance on the CRAG dataset. A summary of these results is shown in Table \ref{table:accuracy_results}

\begin{figure}[hbt!]
\centering
\begin{lstlisting}[language=Python]
prompt = f"""
You are a generous evaluator. Your goal is to determine whether the **predicted answer reflects the intended meaning** of the expected answer, even if the wording is different or the match is only partial.

- You should be lenient: allow paraphrases, synonyms, and generalizations.
- The predicted answer does not need to match all details, just the key idea.
- If the main meaning of the expected answer appears clearly or recognizably in the predicted answer-even in different words-consider it a match.
- Do not accept answers that are too vague or only weakly related.
- Ignore unrelated or repetitive sentences.

Query:
{row[queryCol]}

Expected answer (main idea to be reflected):
{row['expected']}

Predicted Answer:
{row[predictedCol]}

Does the predicted answer contain the **main idea or intent** of the expected answer in any form (direct, indirect, partial, or paraphrased)?

Respond only with "Yes" or "No".
"""
\end{lstlisting}
\caption{Prompt used for LLM-as-a-judge}
\label{lst:llm_evaluator}
\end{figure}

\begin{table}[!htbp]
\centering
\caption{Accuracy of KG-based QA System under Different Evaluation Settings}
\label{table:accuracy_results}
\renewcommand{\arraystretch}{1.5}
\begin{tabular}{| p{9cm} | p{1.5cm} | p{1.5cm} |}
\hline
\textbf{Questions Asked} & \textbf{Llama 3.2 Accuracy} & \textbf{GPT-3.5 Turbo Accuracy} \\
\hline
QAs generated at runtime using Hugging Face's model (set of 194) & 89.6\% & 78.3\% \\
QAs generated at runtime using Hugging Face's model and later perturbed (set of 194) & 85.6\% & 72.6\% \\
QAs from standard CRAG dataset (set of 2700) & 68.6\% & 50\% \\
\hline
\end{tabular}
\end{table}

\subsection{Robustness Evaluation via Question Perturbation}
To assess the robustness and generalization capabilities of the system, the questions were perturbed while preserving their original intent using Llama 3.2. The modified questions were then re-evaluated to determine whether the system could still retrieve correct answers. One of the examples of the perturned QA is shown in Fig. \ref{lst:perturb_sample}. The accuracy of the perturbed questions on the PDF dataset is 85.6\% with Llama-3.2 and 72.6\% with gpt-3.5-turbo.

\begin{figure}
\centering
\begin{lstlisting}
Original Question: Which of the following is NOT an Employer's Risk under Clause 3.1?
Answer: Contractor's Negligence is not the employer's risk.

Perturbed Question: Which is not the responsibility of the employer according to Clause 3.1?
Answer: Contractor's Negligence is not the employer's risk. (Correct)
\end{lstlisting}
\caption{Perturbation Sample}
\label{lst:perturb_sample}
\end{figure}

\section{CONCLUSION}
This research demonstrates a knowledge graph-based approach for question answering. The pipeline combines entity and semantic search with reranking and paraphrasing. The system is practically useful as it provides interpretable, ranked responses that include partially correct or contextually relevant information, making it valuable for exploratory QA, entity discovery, and knowledge validation. Future directions include fine-tuned QA pair generation and richer relation extraction. The code for the experiment, methods, and dataset described in this paper is publicly available at \url{https://github.com/kartikeyaaneja/GraphRAGDirectAnswer}.